# Revisiting the Hierarchical Multiscale LSTM

Ákos Kádár
Tilburg University
a.kadar@uvt.nl

Marc-Alexandre Côté
Microsoft Research Montreal
macote@microsoft.com

Grzegorz Chrupała
Tilburg University
g.chrupala@uvt.nl

Afra Alishahi
Tilburg University
a.alishahi@uvt.nl

## Abstract

Hierarchical Multiscale LSTM (Chung et al., 2016a) is a state-of-the-art language model that learns interpretable structure from character-level input. Such models can provide fertile ground for (cognitive) computational linguistics studies. However, the high complexity of the architecture, training procedure and implementations might hinder its applicability. We provide a detailed reproduction and ablation study of the architecture, shedding light on some of the potential caveats of re-purposing complex deep-learning architectures. We further show that simplifying certain aspects of the architecture can in fact improve its performance. We also investigate the linguistic units (segments) learned by various levels of the model, and argue that their quality does not correlate with the overall performance of the model on language modeling.

## 1 Introduction

Verifying and reproducing claims published in scientific articles is an essential part of building a solid foundation for future research. As such, reproduction has a long history in many scientific fields (Willett et al., 1985; Venables et al., 1993; Waltemath et al., 2011). Recent large-scale studies, however, raise concerns about the reproducibility in a variety of areas and the potential effect of this crisis (Baker, 2016). A reproduction study by the Open Science Collaboration (2015) estimates that only 40% of research in psychology is reproducible, while Begley and Ellis (2012) end up confirming only 11% of preclinical cancer studies. The latter work makes an important link between the low reproducibility rates and the notoriously low impact of preclinical cancer research on clinical practice (Hutchinson and Kirk, 2011).

Our work is motivated by a similar concern, specifically the applicability of complex deep learning architectures for computational (cognitive) linguistics research. State of the art systems often employ complex architectures which integrate various design features and use many optimization techniques. Because the focus is on boosting the final performance on a given task, often little effort is put into understanding where the power of the system comes from. This makes these models much harder to adapt for new tasks or domains. In our view, it is essential not only to be able to reproduce reported results, but also to understand the contribution of various design features through systematic ablation experiments.

The higher performance brought by modern neural network architectures often comes at the cost of our understanding of the representations and structural information the system learns. However, for models to be generalizable to new domains, it is important to move towards analyzing such structural representations, and investigating their impact on the final performance of the model.

In the current study, we examine the reproducibility of a language model with the ability to learn explicit linguistic structure: the Hierarchical Multiscale Recurrent Neural Network (HMLSTM) model. This architecture was introduced by Chung et al. (2016a) and set a new state of the art on language-modeling benchmarks Text8, Hutter Prize and character-level Penn Treebank. Additionally, the paper features examples where the lowest layer of the model recovers word-level segmentation and in some cases performs interpretable morphological analysis. With this reproduction study our motivation is to provide insight into the dynamics of the model, which can facilitate re-using and further developing its architecture and repurposing it for tasks other than language modeling. Crucially, we investigate whether the performance of the model depends on the acquisition of high-quality linguistic structure.

### 1.1 The importance of reproducibility

There is no established tradition of reproduction studies in computational linguistics; however, scattered attempts at reproducing a number of studies have highlighted the importance of investigating the dataset, the model architecture and the evaluation scheme used in experimental designs. For example, Mieskes (2017) quantifies the availability of non-benchmark datasets underlying the experiments. Horsmann and Zesch (2017) re-run coarse-grained multilingual part-of-speech tagging experiments by Plank et al. (2016) and confirm the superior performance of LSTM-based architectures on fine-grained tagsets. Marrese-Taylor and Matsuo (2017) fail to reproduce the results of three articles in the domain of aspect-based opinion mining, with the conclusion that repeating experiments without the availability of source code is hindered due to lack of details on pre-processing, model architecture specification and exact parameter settings. Morey et al. (2017) replicate the results of 9 discourse parsers trained on the RST Discourse Treebank (RST-DT) (Carlson et al., 2003) and show that most of the recent gains in the domain are due to non-trivial differences in evaluation methodology.

We propose one further step in this direction: in addition to reproducing reported results, it is important to investigate the role the components of complex models play and examine their impact on the behavior of the model. The HMLSTM model that we choose as our case study has many desirable properties. However, it is fairly complex and allows for a considerable degree of freedom for implementation. We experimentally explore the function of architectural details of the HMLSTM model, as well as experiment with varying the specifics of the training procedure. We aim to shed light on the potential bottlenecks involved in re-purposing a complex deep learning sequence modeling architecture for computational linguistics studies.

### 1.2 The importance of interpretability

Language data, both in the form of speech and text, exhibits hierarchical structure: characters or phonemes form morphemes and words, which form phrases, which in turn form whole sentences. At each level, the units are composed to build up the meaning of the sentence at the highest level. Modern RNN architectures have been very successful in solving character-level NLP tasks (Chung et al., 2016b; Golub and He, 2016; Mikolov et al., 2012). However, they do not make the learned linguistic structure explicit: rather it can be presumed to be cryptically encoded in the states of the hidden layers. The higher performance brought by modern neural network architectures often comes at the cost of our understanding of the representations and structural information such systems learn. However, for a model to be generalizable to new domains, it is important to move towards analyzing such structural representations, and investigating their impact on the final performance of the model.

Understanding the structural information encoded in black-box neural network architectures has been a desired target since Elman (1990) and several studies have used indirect post-analysis techniques and auxiliary tasks to probe the acquired structure. For example, recent studies have assessed the ability of Long Short-Term Memory (LSTM) language models to encode subject-verb agreement (Linzen et al., 2016), analyzed the translation quality of character-level sequence-to-sequence models in terms of several morpho-syntactic grammaticality tests (Sennrich, 2016), Li et al. (2016) introduce a representation erasure technique to measure the amount of contribution of input words and specific phrases to the decision of RNN-based sentiment classifiers, Kádár et al. (2017) examine the linguistic representations of Gated Recurrent Unit-based architectures for image-sentence ranking, and Chrupała et al. (2017) and Alishahi et al. (2017) analyze the linguistic structure learned by Recurrent Highway Network models of visually grounded speech understanding.

As an alternative approach, a number of *white-box* architectures have been proposed which learn explicit representations of linguistic structure. For example, Dyer et al. (2016) examine Recurrent Neural Network grammars and analyze its learned syntax (Kuncoro et al., 2016). They conclude that their architecture learns a similar notion of headedness as established head-rule sets and the model learns structure similar to traditional nonterminal categories. Williams et al. (2017) explore the learned grammars of two state-of-the-art natural language inference models: SPINN (Bowman et al., 2016) and Gumbel Tree-LSTM (Choi et al., 2017). They find that both SPINN and Gumbel Tree-LSTM have close to or worse than chance-level parsing performance on standard benchmarks. Furthermore, they note that in the case of the SPINN architecture the learned structure depends on the tuning of the model and that their findings are different from the SPINN implementation of Yogatama et al. (2016). They further note that consistency of the parses produced by Gumbel Tree-LSTM across multiple runs is not far off chance level.

The Hierarchical Multi-scale LSTM architecture (Chung et al., 2016a) is in the latter class of white-box models. It is designed specifically to allow for learning explicit structure in data: at each layer, it predicts a binary boundary variable at each time-step. These boundaries correspond to an explicit, interpretable segmentation of the input utterance at multiple levels, which can be thought of as different levels of granularity in linguistic structure. At the same time, it reportedly has state-of-the-art performance on several language modeling benchmarks. As such, we believe the model has the potential to impact future computational cognitive linguistics studies. However, it is not clear whether the high performance of this model demands the acquisition of high-quality linguistic structure from input text. Furthermore, due to the complex architectural design, implementation details might play a large role in the conclusion that can be drawn from the learned structure as in Williams et al. (2017). We examine the segmentation of the input characters by different versions of this model to see whether their quality impacts the overall performance.

## 2   Hierarchical Multiscale LSTM

Let us introduce the notations used throughout this section with the standard LSTM equations:

$$\begin{bmatrix} \mathbf{i}_t \\ \mathbf{f}_t \\ \mathbf{u}_t \\ \mathbf{o}_t \end{bmatrix} = W\mathbf{x}_t + U\mathbf{h}_{t-1} + \mathbf{b} \qquad \text{(gates and candidate)}$$

$$\mathbf{c}_t = \mathbf{c}_{t-1} \odot \sigma(\mathbf{f}_t) + \tanh(\mathbf{u}_t) \odot \sigma(\mathbf{i}_t) \qquad \text{(cell state)}$$

$$\mathbf{h}_t = \sigma(\mathbf{o}_t) \odot \tanh(\mathbf{c}_t) \qquad \text{(hidden state)}$$

The input to an LSTM are the current input $\mathbf{x}_t$, previous state $\mathbf{h}_{t-1}$ and previous cell state $\mathbf{c}_{t-1}$. Variables $\mathbf{x}_t$ and $\mathbf{h}_{t-1}$ are used to compute the input gate $\mathbf{i}_t$, forget gate $\mathbf{f}_t$, candidate activation $\mathbf{u}_t$ and output gate $\mathbf{o}_t$. In the second equation $\mathbf{f}_t$ and $\mathbf{i}_t$ are used to trade off the candidate activation $\mathbf{u}_t$ and the previous cell-state $\mathbf{c}_{t-1}$ using the element-wise multiplication $\odot$. This results in the current cell state $\mathbf{c}_t$. Finally, the output gate soft-selects the components of $\tanh(\mathbf{c}_t)$ to write to the hidden state $\mathbf{h}_t$. The function $\sigma$ refers to the sigmoid function.

**HMLSTM –**   To make the explanation of the HMLSTM architecture more straight-forward, let us first consider the bottom- and top-layer as they are special cases of any $\ell$-th layer. In addition, we provide vectorized version equations of the HMLSTM architecture which allow for batch sizes larger than 1 (see Appendix A).

**Bottom Layer –**   The bottom layer of the HMLSTM takes as input: 1) *bottom-up connection*: input at the current time-step $\mathbf{x}_t$; 2) *recurrent connection*: hidden and cell states

$<\mathbf{h^1}_{t-1}, \mathbf{c^1}_{t-1}>$ and previous boundary variable $z^1_{t-1}$ from $\ell = 1$; and 3) *top-down connection*: hidden state $\mathbf{h^2}_{t-1}$ from $\ell = 2$. It outputs the state tuple $<\mathbf{h}^1_t, \mathbf{c}^1_t>$ and boundary variable $z^1_t$.

$$\begin{bmatrix} \mathbf{i}_t \\ \mathbf{f}_t \\ \mathbf{u}_t \\ \mathbf{o}_t \\ z_t \end{bmatrix} = W\mathbf{x}_t + U\mathbf{h}^1_{t-1} + z_{t-1}V\mathbf{h}^2_{t-1} + \mathbf{b} \tag{1}$$

The activation function on $z_t$ is the hard-sigmoid followed by the rounding operation to discretize it to $\{0, 1\}$: $z_t := \text{round}(\text{hard\_sigmoid}(z_t))$. The top-down connection is turned on and off by $z_{t-1}$. When the layer detects a boundary in the previous step (i.e. $z_{t-1} = 1$), takes context from the higher layer through the top-down connection. After computing the activations the layer either runs a regular LSTM UPDATE or does a FLUSH of its input:

$$\mathbf{c}_t = \begin{cases} \mathbf{c}_{t-1} \odot \sigma(\mathbf{f}_t) + \tanh(\mathbf{u}_t) \odot \sigma(\mathbf{i}_t) & \text{if } z_{t-1} = 0 \text{ (UPDATE)} \\ \tanh(\mathbf{u}_t) \odot \sigma(\mathbf{i}_t) & \text{if } z_{t-1} = 1 \text{ (FLUSH)} \end{cases} \tag{2}$$

The final hidden state is given by the regular LSTM update $\mathbf{h}_t = \sigma(\mathbf{o}_t) \odot \tanh(\mathbf{c}_t)$.

**Top Layer** – The top layer has both bottom-up $\mathbf{h}^{\ell-1}_t, z^{\ell-1}_t$ and recurrent connections $<\mathbf{h}^\ell_{t-1}, \mathbf{c}^\ell_{t-1}>$, but no top-down connection. At each step, it outputs the state tuple $<\mathbf{h}^\ell_t, \mathbf{c}^\ell_t>$:

$$\begin{bmatrix} \mathbf{i}_t \\ \mathbf{f}_t \\ \mathbf{u}_t \\ \mathbf{o}_t \end{bmatrix} = z^{\ell-1}_t W\mathbf{h}^{\ell-1}_t + U\mathbf{h}^\ell_{t-1} + b \tag{3}$$

The bottom-up input $\mathbf{h}^{\ell-1}_t$ is gated by $z^{\ell-1}_t$. When $z_t = 1$ the higher layer takes one step and performs UPDATE; otherwise it ignores the current time-step and performs COPY:

$$\mathbf{c}_t = \begin{cases} \mathbf{c}_{t-1} \odot \sigma(\mathbf{f}_t) + \tanh(\mathbf{u}_t) \odot \sigma(\mathbf{i}_t) & \text{if } z_t = 1 \text{ (UPDATE)} \\ \mathbf{c}_{t-1} & \text{if } z_t = 0 \text{ (COPY)} \end{cases} \tag{4}$$

**Middle layers** – The general HMLSTM layer runs all operations: UPDATE, COPY and FLUSH. It takes as input 1) *bottom-up connection*: lower layer state $\mathbf{h}^{\ell-1}_t$ and boundary variable $z^{\ell-1}_t$; 2) *recurrent connection*: hidden and cell states $<\mathbf{h}^\ell_{t-1}, \mathbf{c}^\ell_{t-1}>$ and boundary variable $z^\ell_{t-1}$; and 3) *top-down connection*: hidden state $\mathbf{h}^{\ell+1}_{t-1}$. It outputs the tuple $<\mathbf{h}^\ell_t, \mathbf{c}^\ell_t>$ and boundary variable $z^\ell_t$:

$$\begin{bmatrix} \mathbf{i}_t \\ \mathbf{f}_t \\ \mathbf{u}_t \\ \mathbf{o}_t \\ z_t \end{bmatrix} = z^{\ell-1}_t W\mathbf{h}^{\ell-1}_t + U\mathbf{h}^\ell_{t-1} + z^\ell_{t-1}V\mathbf{h}^{\ell+1}_{t-1} + b \tag{5}$$

The bottom-up connection is masked by $z^{\ell-1}_t$ as in the top-layer, while the top-down connection is masked by $z^\ell_{t-1}$ as in the bottom-layer. The variables $z^{\ell-1}_t$, $z^\ell_{t-1}$ determine which of the three operations the cell runs. FLUSH is executed if the layer detects a boundary in the previous step (i.e. $z^\ell_{t-1} = 1$). If this is not the case and the lower layer does not detect a boundary, the layer runs COPY. However, if the lower layer does detect a boundary $z^{\ell-1}_t$, it runs UPDATE:

$$\mathbf{c}_t = \begin{cases} \tanh(\mathbf{u}_t) \odot \sigma(\mathbf{i}_t) & \text{if } z^l_{t-1} = 1 \text{ (FLUSH)} \\ \mathbf{c}_{t-1} & \text{if } z^\ell_{t-1} = 0 \text{ and } z^{\ell-1}_t = 0 \text{ (COPY)} \\ \mathbf{c}_{t-1} \odot \sigma(\mathbf{f}_t) + \tanh(\mathbf{u}_t) \odot \sigma(\mathbf{i}_t) & \text{if } z^\ell_{t-1} = 0 \text{ and } z^{\ell-1}_t = 1 \text{ (UPDATE)} \end{cases} \tag{6}$$

**Output gate** – The output embedding $\mathbf{h}_t^e$ of the HMLSTM is computed by the following gating mechanism:

$$g_t^\ell = \sigma(\mathbf{w}_\ell^T[\mathbf{h}_1;\mathbf{h}_2;\mathbf{h}_3]) \qquad (7)$$
$$\mathbf{h}_t^e = \text{ReLU}(g_t^\ell W_\ell^e \mathbf{h}_t^\ell) \qquad (8)$$

It takes states from the three layers and computes a gating value for each $g_t^1, g_t^2$ and $g_t^3$ using the parameter vector $\mathbf{w}^\ell$. These scalars are used to weight the $W_l^e \mathbf{h}_t^l$ for each layer where $W_\ell^e$ is a learned parameter. It is important to add some mechanism to combine the activations of the three layers for language modeling as all three layers tick on a different time-scale, but the network has to generate an output at each time-step. This combined embedding $\mathbf{h}_t^e$ is then used in the final softmax layer to predict the probability distribution over the next character.

## 3 Experiments

### 3.1 Experimental setup

Upon our request, the authors of the original paper shared a code base with us implementing the HMLSTM architecture and functionality to run the experiments on Penn Treebank (henceforth PTB). The code was implemented in Tensorflow version 1.1 (Abadi et al., 2016), which we ported with minimal modifications to version 1.6 to be able to run it on our system. In the default settings, the code contains $\ell^2$-norm weights penalty term with a coefficient of 0.0005, which we also kept fixed across experiments. Furthermore, the given source code has a special initialization strategy used for the weights matrices: until the last column vector it is initialized with orthogonal initialization with standard normal distribution. The last column vector (corresponding to the logit of $z$) is initialized with Glorot uniform initialization with $\sigma = \sqrt{\frac{6.0}{\text{fanin}+\text{fanout}}}$. Due to no initialization details in the original paper we used the initialization defaults provided in the code base for all PTB and Text8 experiments. Lastly, the authors did not refer to publicly available pre-processed versions of the datasets and we used our versions following the details of the paper. We made the following changes to the original code:

**Baseline** – The authors report LSTM baselines for both PTB and Text8 data sets. The exact configuration of the baseline is not discussed in the paper, however, the authors do share that they implement both layer-normalization and the output module. We train a three-layer stacked LSTM with the same hyperparamters as the HMLSTM.

**Layer normalization** – The original paper reports results using layer normalization (Ba et al., 2016), which was not implemented in the code base. There are multiple options where to apply it in the case of a LSTM cell. For the HMLSTM cell we implemented it based on equations 20-22 in the paper introducing the technique (Ba et al., 2016). Furthermore, we applied layer-norm to the input and output embedding layers since the performance of the model was found to be unstable otherwise.

**Learning-rate schedule and early-stopping** – The code base did not include an implementation of the learning-rate schedule that was reported to be used in the original experiments. The learning-rate schedule reported in the paper divides the learning rate by 50 when no improvement is observed in the validation loss. In our code we monitor the validation loss at each epoch. The original code base ran the models for 500 epochs with no early-stopping and no learning-rate schedule, however, we use early stopping with patience of 4 as very small improvements are observed after reducing the learning rate four times by 50.

**Evaluation** – We report the bits-per-character (bpc) of the trained models by computing the entropy (in base 2) of the entire test set and dividing it by the number of characters. The code did not implement the evaluation procedure and we have not found the exact details in the paper. Our implementation follows Mikolov et al. (2012): we use batch size of 1 and carry the states of the recurrent model over to the following batch of 100 characters until the whole test sequence is processed.

The experimental results reported in Section 4 are all using this revised implementation, manipulating various ablation factors as described in Section 3.2. Following Chung et al. (2016a), we report results on the following two datasets.

**Character-level Penn Treebank** – The smaller-scale experiments apply variations of the model on the Penn Treebank dataset (Marcus et al., 1993) using the splits and preprocessing from Mikolov et al. (2012). All models are trained with sequence lengths of 100 and batch size of 64. Before each epoch the dataset is randomly cropped to be divisible by 100. The parameters are optimized with Adam (Kingma and Ba, 2014) with initial learning rate of 0.002, which is divided by 50 if no improvement on the validation data is observed after a full epoch. The norm of the gradient is clipped at 1.0. For all models we use 512 units for all layers and 128 dimension character embeddings as reported in the paper.

**Text8** – The Text8 dataset (Mahoney, 2011) is extracted from Wikipedia and is a sequence of 100 million alphabetical characters and spaces. The splits used by Chung et al. (2016a) correspond to the (by now standard) splits introduced in Mikolov et al. (2012): first 90M characters for training, the next 5M for validation and the final 5M characters for testing. On this dataset models are trained on sequence length of 100 and batch size of 128. The initial learning rate of Adam was set to 0.001 and we applied the learning-rate schedule and early-stopping strategy described before. The norm of the gradient is clipped at 1.0. We use 1024 units for the HMLSTM layers and 2048 output embedding units.

Like in the original paper, models are trained to minimze the log-likelihood of the training set and the non-differentiability gap caused by the rounding operation when discretizing the boundary variables is solved by using the straight-through estimator (Bengio et al., 2013).

### 3.2 Ablation factors

In order to analyze the behavior of the model and the impact of its various architectural design features as well as evaluation settings, we manipulate the following factors in our experiments.

**Layer normalization (LN).** As mentioned in the previous section, there was a mismatch between the description of the HMLSTM model in Chung et al. (2016a) and the code base provided to us in terms of layer normalization. Therefore we report our experimental results both with and without layer normalization.

**Learning-rate schedule (Schedule).** Similarly, the paper and the code base do not match in applying learning-rate schedule, therefore we report results with and without scheduling.

**Sensitivity to slope ($\alpha$).** Hard-sigmoid (Gulcehre et al., 2016) is a piecewise linear approximation of the sigmoid function, defined as the first-order Taylor expansion of the sigmoid around $x \approx 0$ and clipped between the limits of the original function (0 and 1): $\max(0, \min(1, 0.25x + 0.5))$; where 0.25 is the slope $\alpha$. The authors describe the slope annealing trick for the PTB experiments and state that they start from slope 0.5 and slowly increase it at maximum until 2.5. We have not found details regarding the $\alpha$ values in other experiments, but used the default $\alpha = 0.5$ in the code-base. We test whether the hard-sigmoid slope $\alpha = 0.25$ reaches the same performance as the specific choice of $\alpha = 0.5$. Furthermore, we gauge sensitivity to $\alpha$ using values 0.125 and 1.0.

**COPY operation on the last layer (CopyLast).** The code base by default, only applies the COPY operation at the last layer on the cell state $c_t^l$, but not on the hidden state $h_t^l$. In the latter case if the layer runs COPY it takes the previous cell state and applies the new output gate on it along with the tanh function $\mathbf{o}_t \odot \tanh(\mathbf{c}_{t-1})$. This is different from the original formulation of the HMLSTM. The main experiments were run with this default, but we also report the impact of changing this parameter.

|  | PTB | Text8 |
|---|---|---|
| HMLSTM reported by (Chung et al., 2016a) | 1.25 | 1.29 |
| 3-layer LSTM reported by (Chung et al., 2016a) | 1.29 | 1.39 |
| 3-layer LSTM + Schedule + LN | 1.32 | 1.37 |
| HMLSTM + Schedule + LN | 1.27 | 1.36 |
| HMLSTM + Schedule + LN + CopyLast | 1.29 | 1.36 |

Table 1: Reproducing language modeling results (in bpc) on PTB and Text8 datasets.

**Top-down connection (NoTopDown).** When the HMLSTM cell runs the FLUSH operation, it computes the new cell-state by $\mathbf{i}_t \odot \mathbf{u}_t$ and the previous cell-state is dropped from the computation. However, both $\mathbf{i}_t$ and $\mathbf{u}_t$ depend on the previous state $\mathbf{h}_{t-1}$. As such, information from the past steps still "leaks" even at FLUSH and each layer can potentially retain enough information for future computations without the use of top-down connections. In our experiments we run the HMLSTM model with and without top-down input to evaluate the impact of this simplification.

**Simpler output layer (SimplerOut).** The hidden state of the last layer of the HMLSTM is not updated at every time-step and as such it computes a gated embedding combining the representation of all three layers to be able to provide a prediction for every time-step. Here we implement a simplification to the output layer by replacing it with a single affine transformation, embedding the concatenated hidden states: $\mathbf{h}_t^e = \text{ReLU}(W^e[\mathbf{h}_1; \mathbf{h}_2; \mathbf{h}_3])$.

**Alternative architecture (HMRNN).** One of the motivations for the HMLSTM architecture in the original paper (Chung et al., 2016a) is that the hierarchical multiscale structure might help with the vanishing gradient problem as gradients are back-propagated through fewer time-steps. Here we introduce an alternative architecture that we call HMRNN, which replaces the LSTM updates with simple Elman RNN recurrence. It is a much simpler model and it helps in understanding if the hierarchical multiscale structure itself is enough to alleviate the vanishing gradient problem, or if the LSTM-style updates are essential for the good performance.

$$\mathbf{h}_t = \begin{cases} \tanh(W\mathbf{h}_t^{l-1} + V\mathbf{h}_{t-1}^{l+1}) & \text{if } z_{t-1}^l = 1 \text{ (FLUSH)} \\ \mathbf{h}_{t-1} & \text{if } z_{t-1}^l = 0 \text{ and } z_t^{l-1} = 0 \text{ (COPY)} \\ \tanh(W\mathbf{h}_t^{l-1} + U\mathbf{h}_{t-1}^l) & \text{if } z_{t-1}^l = 0 \text{ and } z_t^{l-1} = 1 \text{ (UPDATE)} \end{cases} \quad (9)$$

The HMRNN architecture implements a simple formulation of the hierarchical multiscale intuition. In case of the FLUSH operation, it only takes as input the hidden state of the lower layer and the hidden state of the higher layer at the previous step. In case of COPY it just uses the previous hidden state. Finally, when running UPDATE it applies the simple Elman network update. See Appendix B for a vectorized implementation.

## 4 Experimental Results

### 4.1 Reproducing language modeling results

We attempt to replicate the language modeling experiments of Chung et al. (2016a) on the character level Penn Treebank and Text8 datasets. Table 1 shows the language modeling results reported by Chung et al. (2016a) on both datasets. It also reports results of our reproduction of the original experiments as described in the paper. This means using the HMLSTM architecture with layer normalization on all layers, and with learning-rate scheduling. Since the use of the COPY operation on the last layer is ambiguous and not consistent between the original paper and the code base, we report results both with COPY on and off.

Our most faithful reproduction of the original experiments, which uses the HMLSTM architecture with learning-rate schedule, layer normalization and COPY on the last layer, results in 1.29 bpc on the PTB dataset. Dropping the COPY option improves the results to 1.27 bpc, which is still below the results reported in Chung et al. (2016a) on the same dataset (1.25). Our 3-layer LSTM baseline on PTB also under performs the baseline reported by Chung et al. (2016a)(1.32 compared to 1.29 bpc).

On the Text8 dataset, our results (1.36) are much further from the published ones (1.29), with the HMLSTM performing very close to the 3-layer LSTM baseline (1.37). We did not pursue further experiments on this dataset. The discrepancy could be due to the fact that the script that the authors shared with us was optimized for PTB, and some of the settings might need to be changed for Text8. However, since we have not found such differences reported in the paper, we used the same setting for both datasets. Interestingly, contrary to the HMLSTM results, our implementation of the 3-layer LSTM basline improves the reported baseline performance from 1.39 to 1.37 bpc.

## 4.2 Ablation results

|    |                                              | BPC  | Iter. | $z^1$ | $z^2$ | z-ratio |
|----|----------------------------------------------|------|-------|-------|-------|---------|
| 1  | HMLSTM + Schedule + LN + copylast            | 1.29 | 10K   | 0.41  | 0.25  | 1.64    |
| 2  | HMLSTM                                       | 1.39 | 16K   | 0.23  | 0.15  | 1.53    |
| 3  | HMLSTM + Schedule                            | 1.32 | 12K   | 0.25  | 0.16  | 1.56    |
| 4  | HMLSTM + Schedule + LN                       | 1.27 | 12K   | 0.42  | 0.09  | 4.67    |
| 5  | HMLSTM + Schedule + LN + $\alpha$=0.125      | 1.28 | 14K   | 0.41  | 0.23  | 1.78    |
| 6  | HMLSTM + Schedule + LN + $\alpha$=0.25       | 1.27 | 12K   | 0.31  | 0.21  | 1.48    |
| 7  | HMLSTM + Schedule + LN + $\alpha$=1.0        | 4.32 | 4K    | 1.0   | 1.0   | 1.0     |
| 8  | NoTopDown + Schedule + LN                    | 1.28 | 12K   | 0.64  | 0.28  | 2.29    |
| 9  | SimplerOutput + Schedule + LN                | **1.25** | 10K | 0.64  | 0.31  | 2.06    |
| 10 | 3-layer LSTM + Schedule + LN                 | 1.32 | 9K    | N/A   | N/A   | N/A     |
| 11 | 3-layer LSTM + Schedule + LN + SimplerOut    | 1.26 | 12K   | N/A   | N/A   | N/A     |
| 12 | HMRNN + Schedule + LN                        | 1.40 | 18K   | 0.0   | 1.0   | 0.0     |

Table 2: Ablation results on PTB

Since we did not succeed in reproducing the original results on Text8, we only perform the ablation experiments on the PTB dataset. Table 2 summarizes the results of these experiments. On the first row, we repeat our reproduction results using the most faithful replication of the original experiments (HMLSTM + Schedule + LN + CopyLast). We will compare the results from various modifications of the architecture or evaluation setup to these base results.

**Layer normalization and learning-rate schedule –** As is evident from the results on rows 2–4, adding learning-rate schedule and layer normalization improve the results over the base HMLSTM. Furthermore, the results on row 4 (HMLSTM + Schedule + LN) are better than the reproduction results which also implement the CopyLast feature. Therefore, in the rest of the experiments, we consistently add layer normalization on all layers as well as learning-rate scheduling, and drop CopyLast.

**Sensitivity to alpha –** Rows 5–7 show results for different values of the slope $\alpha$. Using the by definition hard-sigmoid slope $\alpha = 0.25$ instead of the default $\alpha = 0.5$ in the code does not degrade the performance. Lowering alpha to $\alpha = 0.125$ degrades the performance slightly. Since

Chung et al. (2016a) mention that they increase the slope up to $\alpha = 2.5$, we also tested the results with a higher value of $\alpha = 1.0$ (row 7), which caused the model to diverge after 4,000 iterations.

**Architectural modification: top-down connection and simpler output –** Interestingly, as can be seen from row 8, removing the top-down connections (and keeping the model structure fixed otherwise) only degrades the performance by a small margin; from 1.27 bpc to 1.28. Even more surprising is the effect of using a simpler output layer: replacing the gated-output layer by our simple output module improves the results from 1.27 to 1.25. This, in fact, is the best results we have achieved. Similarly rows 10 and 11 show that the simpler output layer improves the baseline 3-layer LSTM performance from 1.32 bpc to a competitive 1.26 bpc.

**Alternative architecture: HMRNN –** The simplified architecture which replaces the LSTM updates with Elman RNN produces results that are much worse (1.40) than the HMLSTM counterpart (1.27). Our ablation experiments show that using layer normalization and learning-rate schedule is beneficial, high values for the slope of the sigmoid function hurt performance, but the model performance is robust otherwise to smaller $\alpha$ values. The LSTM updates are essential for good performance of the HMLSTM model. Finally, simplifying certain aspects of the model such as removing the top-down connection might not hurt much, and in other cases such as simplifying the output layer it can actually enhance the overall performance.

### 4.3 Segmentation results

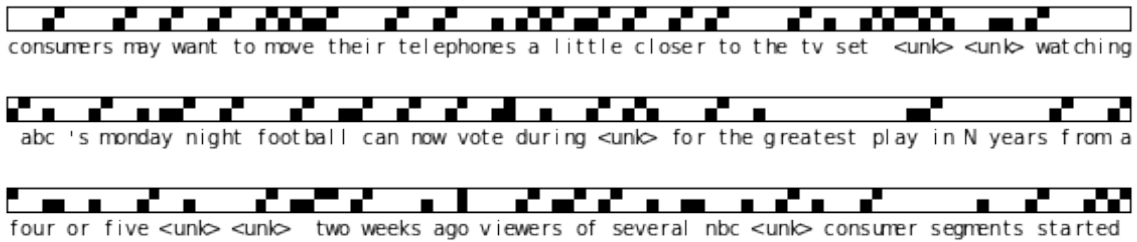

Figure 1: Segmentation examples. Black means $z_t = 1$, white $z_t = 0$.

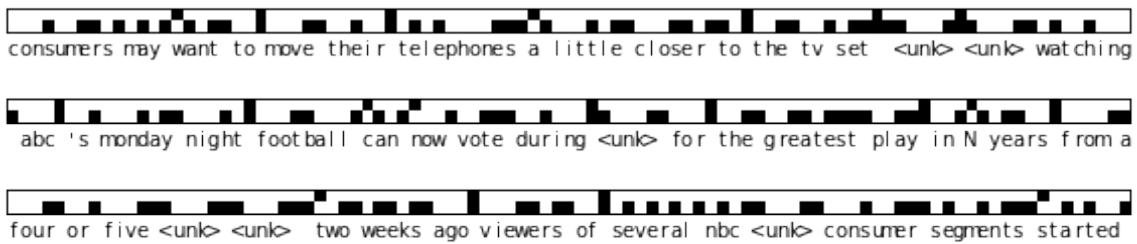

Figure 2: Segmentation examples with layer normalization. Black means $z_t = 1$, white $z_t = 0$.

The HMLSTM model predicts binary boundary variables $z_t^1$ and $z_t^2$ at each time-step, which correspond to a segment of the input utterance at each level. The frequencies of $z_t^1$ and $z_t^2$ provide a high-level description of segmentation properties of the network and are reported in Table 2. Based on the hierarchical multiscale intuition, the expected behavior is that the $z_t^1$ frequency is higher than $z_t^2$ as the former corresponds to words or morpheme-like units, whereas the latter corresponds to larger chunks. The ratio of $z_t^1$ to $z_t^2$ frequencies is reported in the last column of Table 2.

Running the HMLSTM default configuration with added layer-normalization and learning-rate schedule results in the largest separation in the time-scale of the layers as indicated by the $z$-ratio of 4.67. Interestingly, only changing the $\alpha$ value to 0.125 or 0.25 results in a much narrower gap (1.78 and 1.48 respectively), without a big impact on bpc. Furthermore, the HMLSTM with schedule, layer-norm and $\alpha = 0.125$ in row 5 performs on the same level as our NoTopDown ablation architecture, with a smaller $z$-ratio.

The segmentation examples reported in the original paper show the first layer of the HMLSTM segmenting the sequence at word boundaries (i.e. spaces), and the higher layer segments corresponding to multi-word chunks. Figure 1 shows segmentation results of the HMLSTM with learning-rate schedule and Figure 2 with the added layer normalization. Our runs could not reproduce the segmentation results visualized in the original paper and overall we do not find a relationship between the performance of the model and segmentation behavior.

## 5 Conclusion

In our attempt to reproduce the results in Chung et al. (2016a) we re-ran the HMLSTM model on two datasets: PTB and Text8. On PTB the final performance of the model almost matches the original results, but not quite. Our best result was achieved by a slight modification to the COPY operation of the last-layer provided in the source-code, but not detailed in the paper and with our added simplification to the output layer.

We could not reproduce the results on Text8 using the same code base received from the authors. This might be due to the non-availability of the pre-processed version of the dataset. Another potential source might be the lack of detail in the paper on the initialization scheme and weight penalty, which led us to keep these implementation details constant across datasets. Furthermore, the layer normalization might have been implemented on different layers across different datasets. We have made several attempts until reaching the conclusion that it needs to be implemented on every layer of the HMLSTM. This was found to perform best on PTB, but might not be the best setting for Text8. Similarly, the learning-rate schedule was applied based on the monitoring of the loss after every epoch, but this was an informed guess.

Two simplifications were applied to the architecture successfully: 1) removing top-down connections only slightly degraded performance, and 2) simplifying the output layer improved performance in our experiments. There is space for possible modifications that fell out of the scope of the current work like applying REINFORCE gradients in place of the straight-through estimator.

We could not reproduce the segmentation results provided in the paper. Furthermore, we did not observe a close relationship between the segmentation behavior and the final performance. Changing the slope variable $\alpha$ to 0.25 from 0.5, while keeping all other details constant, resulted in the same performance, but huge difference in segmentation behavior.

The observations made here are not specific to the work of Chung et al. (2016a) and the HMLSTM architecture. Similar results have been reported recently by Williams et al. (2017), whose re-implementation of the considered architecture did not produce the reported performances. One of their considered models learned a qualitatively different latent structure than another re-implementation by Yogatama et al. (2016) and the other architecture did not converge to the same structure across runs. The HMLSTM architecture was considered for reproduction due to its intriguing property of learning interpretable structure from character-level input. The detailed reproduction and ablation study was provided to surface some of the potential difficulties that can hinder the applicability of such models for future studies.

## A  Vectorized formulation of the HMLSTM

Here are the vectorized equations for the bottom, middle and top layers of the HMLSTM. Note, the computation process should be executed from top to bottom and left to right when there are multiple equations.

**Bottom layer**
$$\mathbf{c}_t = (1 - z_{t-1})\mathbf{c}_{t-1} \odot \sigma(\mathbf{f}_t) + \tanh(\mathbf{u}_t) \odot \sigma(\mathbf{i}_t) \tag{10}$$

**Top layer**
$$\begin{aligned}
\hat{\mathbf{c}}_t &:= \mathbf{c}_{t-1} \odot \sigma(\mathbf{f}_t) + \tanh(\mathbf{u}_t) \odot \sigma(\mathbf{i}_t) & \hat{\mathbf{h}}_t &:= \sigma(\mathbf{o}_t) \odot \tanh(\mathbf{c}_t) \\
\mathbf{c}_t &:= z_t \hat{\mathbf{c}}_t + (1 - z_t)\mathbf{c}_{t-1} & \mathbf{h}_t &:= z_t \hat{\mathbf{h}}_t + (1 - z_t) \odot \mathbf{h}_{t-1}
\end{aligned} \tag{11}$$

**Middle layer**
$$\begin{aligned}
c_m &= (1 - z_{t-1}^{\ell})(1 - z_t^{\ell-1}) & \mathbf{c}_t &= \mathbf{u}_g + c_m(\mathbf{c}_{t-1} - \mathbf{u}_g) + u_m(\sigma(\mathbf{f_t}) \odot \mathbf{c}_{t-1})) \\
u_m &= (1 - z_{t-1}^{\ell})z_t^{\ell-1} & \mathbf{h}_t &:= \sigma(\mathbf{o}_t) \odot \tanh(\mathbf{c}_t) \\
\mathbf{u}_g &= \sigma(\mathbf{i_t}) \odot \tanh(\mathbf{u_t}) & \mathbf{h}_t &:= c_m \mathbf{h}_t + c_m \odot \mathbf{h}_{t-1}
\end{aligned} \tag{12}$$

Variables $c_m, u_m$ and $\mathbf{u}_g$ stand for copy-mask, update-mask and gated-candidate activation respectively.

## B  Vectorized formulation of the HMRNN

Here are the vectorized equations for the bottom, middle and top layers of the HMRNN. Note, the computation process should be executed from top to bottom and left to right when there are multiple equations.

**Bottom layer**
$$\mathbf{h}_t = \tanh(W\mathbf{x}_t + (1 - z_{t-1}^l)U\mathbf{h}_{t-1}^l + z_{t-1}^l \times V\mathbf{h}_{t-1}^{l+1}) \tag{13}$$

**Top layer**
$$\mathbf{h}_t = (1 - z_t^{l-1}) \times \mathbf{h}_{t-1} + z_t^{l-1} \times \tanh(W\mathbf{x}_t + U\mathbf{h}_{t-1}^l) \tag{14}$$

**Middle layer**
$$\begin{aligned}
c_m &= (1 - z_{t-1}^{\ell})(1 - z_t^{\ell-1}) & h_f &= W\mathbf{h}_t^{l-1} + V\mathbf{h}_{t-1}^{l+1} \\
u_m &= (1 - z_{t-1}^{\ell})z_t^{\ell-1} & h_u &= W\mathbf{h}_t^{l-1} + U\mathbf{h}_{t-1}^l \\
\mathbf{h}_t &= c_m \mathbf{h}_{t-1} + (1 - c_m)\tanh((1 - u_m)h_f + u_m h_u)
\end{aligned} \tag{15}$$

Variables $c_m$ and $u_m$ stand for copy-mask and update-mask activation respectively.